\documentclass[letterpaper, 10 pt, conference]{ieeeconf}  
\IEEEoverridecommandlockouts                              
\usepackage{times}

\usepackage{multicol}
\usepackage[bookmarks=true]{hyperref}

\usepackage[dvipsnames]{xcolor}
\usepackage{graphicx}
\usepackage{cite}
\usepackage{graphics} 
\usepackage{epsfig} 
\usepackage{times} 
\usepackage{amsmath} 
\usepackage{amssymb}  
\usepackage{bm}
\usepackage{subcaption}
\usepackage[ruled]{algorithm2e}
\usepackage{siunitx} 
\usepackage{booktabs} 
\usepackage{makecell}
\usepackage{multirow}
\usepackage{color, colortbl}
\usepackage{adjustbox}
\usepackage{esvect}
\usepackage{tikz}
\usepackage{cuted}

\DeclareMathOperator*{\argmin}{argmin}

\definecolor{Gray}{gray}{0.9}

\newcommand{\mb}{\mathbf}
\newcommand{\etal}{\textit{et al}. }
\graphicspath{{fig/}}
\title{\LARGE \bf Active Mass Distribution Estimation from Tactile Feedback}
\author{Jiacheng Yuan$^{1}$, Changhyun Choi$^{1}$, Ellad B. Tadmor$^{2}$ and Volkan Isler$^{3}$
\thanks{$^{1}$ are with Department of Electrical and Computer Engineering,
        University of Minnesota, Minneapolis, MN, 55455, USA
        {\tt\small yuanx320, cchoi@umn.edu}}%
\thanks{$^{2}$ is with the Department of Aerospace Engineering and Mechanics, University of Minnesota, Minneapolis, MN, 55455, USA
        {\tt\small tadmor@umn.edu}}%
\thanks{$^{3}$ is with the Department of Computer Science and Engineering, University of Minnesota, Minneapolis, MN, 55455, USA
        {\tt\small isler@umn.edu}}%
}

\begin{document}

\maketitle
\thispagestyle{empty}
\pagestyle{empty}

\section{Introduction}
\begin{figure}[t!]
	\centering
	\includegraphics[width=0.42\textwidth]{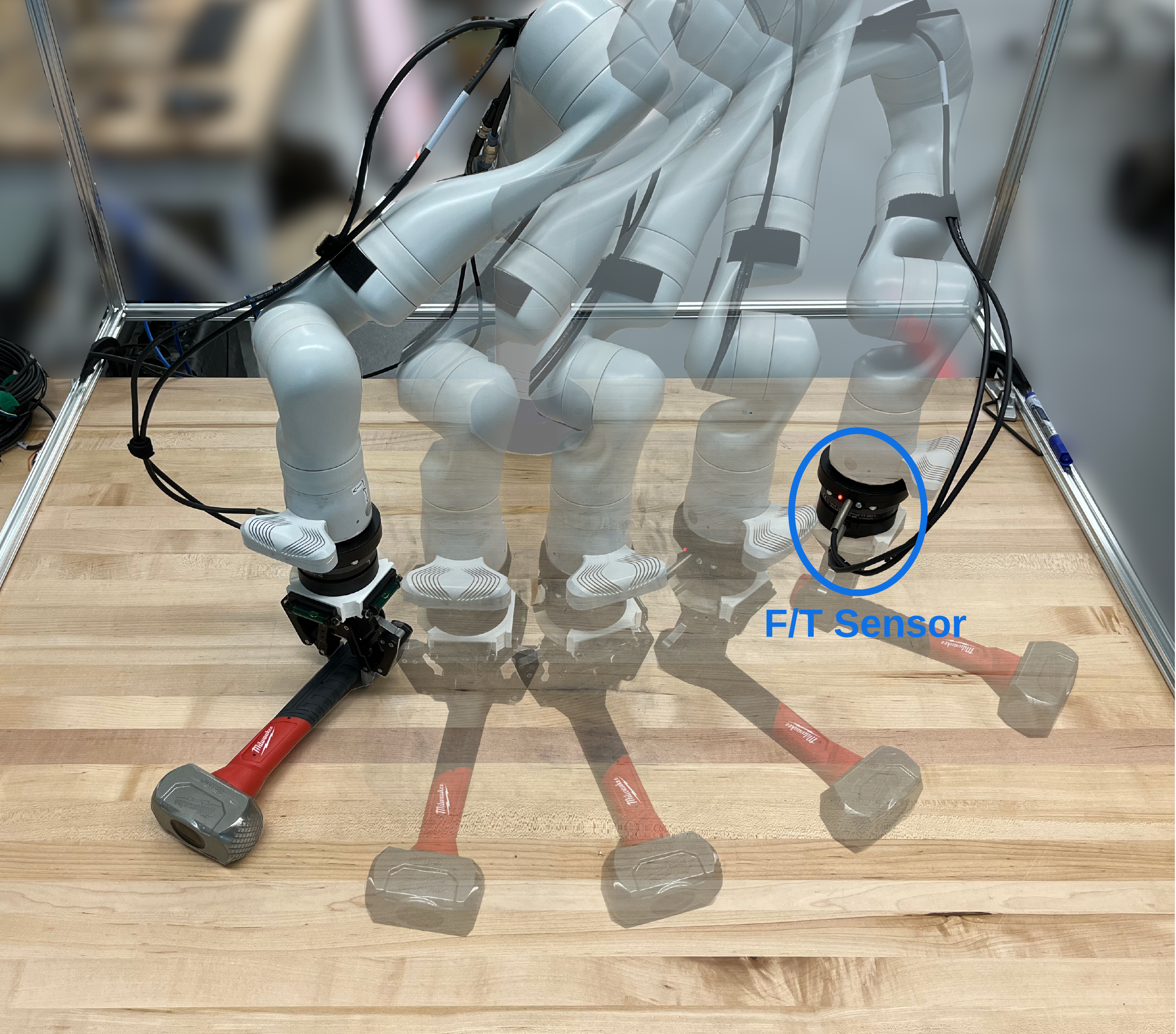}
	\caption{\small{The robot manipulator actively interacts with a rigid object (hammer) on the tabletop by swinging while grasping it. The 6-axis force/torque sensor (marked by a blue ellipse) on the wrist provides tactile feedback from the interaction.}}
	\label{fig:concept}
	\vspace{-6mm}
\end{figure}

In order to manipulate an object or to use it as a tool, it is important to estimate its mechanical properties such as its mass distribution.
Take the hammer in Fig.~\ref{fig:concept} as an example: the center of mass of the hammer, which is critical for grasping it stably, depends on its shape and the specific material it is made of.
The mass distribution also affects the friction interactions between the hammer and the tabletop, which acts on the dynamic behavior of the hammer during non-prehensile manipulation.

Previous studies have shown that robots can estimate such mechanical properties of objects through feedback from various sensor modalities including vision~\cite{martin2016integrated}, audio~\cite{torres2005tapping}, or touch~\cite{kim2014exploration}. Object properties such as shape and pose can be measured without the need for physical interaction with the object~\cite{li2020category, mo2021where2act, standley2017image2mass}. However, mechanical properties usually require forceful interactions and tactile or force feedback measurements. The object mass distribution is no exception. In this work, we focus on how to actively estimate the mass distribution of a rigid object through tactile feedback during object manipulation.

There are a few unique challenges involved in this task. For one, the friction interaction is determined by contact parameters such as contact pressure and friction coefficient. These parameters are usually unknown which complicates the problem. 
Second, it is not clear whether there are actions that excite the unknown parameters. If they exist, how can we choose robot actions that are more informative?  
Lastly, modeling the dynamics of both the robot end-effector and the object means dealing with rigid object contact modeling, which requires complex approximation  methods such as Linear Complementarity Problem [LCP].  
These approximation methods might introduce artifacts to estimation results as well.

To overcome these challenges, we choose to actively apply mechanical perturbations to an object by first grasping it. Thus we restrict the dynamics modeling to only the object itself and its sliding friction interaction with the tabletop. We then adopt a particle-based representation and a differentiable discrete-time dynamics model to describe the physical process where the tactile feedback acts as the control input. Through the analytical formulation, we show how to sequentially estimate the \emph{hidden state }parameters  which can then be used to select robot actions to efficiently estimate the object mass distribution.

Our main contributions are summarized as follows:
\begin{itemize}
	\item We show how to conduct active object mass distribution estimation from tactile feedback on a real robot setup.
	\item We design a strategy to select actions to excite various parameters based on the analytical formulation.
	\item We evaluate our approach against multiple baselines and show that ours outperforms them by a large margin. 
\end{itemize}

\section{Related Work}
We now present an overview of related work.
System identification \cite{swevers1997optimal}  can benefit robotic manipulation by providing information on the object dynamics model. Researchers have studied various tasks that require estimating object properties and physical states during manipulation~\cite{bayrleithner1994static, koval2015pose}.
In \cite{pushnet}, the authors proposed to estimate an object’s center of mass by tracking a history of push interactions with an LSTM module and training an auxiliary objective function. On the other hand, object property parameters can also be incorporated into the dynamics model and be estimated accordingly. Zhang \etal\cite{pf_tactile} showed that by adopting a probabilistic model, tactile feedback can be used to overcome occlusion to perform object pose estimation.  Gaussian Process is one of the widely used approaches to model unknown object dynamics \cite{gp_1, gp_2, gp_3}. Additionally, Model-based Reinforcement Learning can also be applied to learn the dynamics from observing the state action pairs \cite{mbrl_1, mbrl_2, mbrl_3}. 
Works have also been proposed to estimate the object by either IMU \cite{stiffness_1} or pure force sensors \cite{stiffness_2}. 

However, choosing the estimation method is only part of the problem. Action selection also affects the efficiency of the algorithms. A good action will reveal as much information as possible, while poorly selected actions may not excite the target parameters at all. In \cite{van2012maximally}, it is shown that the robot can efficiently infer the composite structure of its environment by quantifying the expected informativeness of its own actions. In \cite{velcro}, authors show that an optimal action selection policy could obtain more information about the environment with uncertainties and bring the system to the manipulation goal more efficiently.
In our work, we analyze the dynamics model and present an efficient strategy to actively select more informative actions.

In robotic manipulation on the tabletop, the interactions between the robot and objects or between objects and the table require modeling for the contact behavior of rigid bodies. 
Bauza \etal\cite{pushing_francois} shown that it is necessary to model the sliding or sticky contact for a planar pushing controller.
Adding these components could further complicate the model. Approximation algorithms such as LCP have been proposed to handle the contact modeling between rigid bodies~\cite{cline2002rigid} and later implemented in auto-differentiation frameworks as a differentiable physics engine in~\cite{end2end}. The differentiable LCP model was used to compute the gradient of the distance between the prediction and the actual observations, and the gradient is then applied to search for values of the model parameters to reduce the reality gap. The same approach was later adopted by Song \etal\cite{song2020identifying} to estimate object mass distribution with random pushing actions. In our work, we show that LCP as an approximation can only close the reality gap to a certain extent, and action selection rules based on the analytical formulation are more efficient for parameter estimation. 

Besides differentiable LCP, many research other works also focus on differentiation through the dynamics model. Degrave \etal \cite{degrave2019differentiable} proposed a differentiable physics engine based on impulse-based velocity stepping methods~\cite{impulse_based_sim} to differentiate control parameters in robotics applications.
In~\cite{combine_model_and_network}, the authors combined analytical dynamics models and neural networks and used end-to-end training to predict the effects of robot actions. New neural network structures were also proposed to directly implement fluid dynamics inside a deep network in \cite{schenck2018spnets} and \cite{liyunzhu_learning}. Differentiable physics was also applied to solve the problem of sequential manipulation and tool-use planning in domains that include physical interactions such as hitting and throwing \cite{toussaint2018differentiable}. In our work, we also implemented the dynamics model as a differentiable physic engine using Taichi Lang \cite{hu2019difftaichi}.

\section{Background}
In this section, we introduce our assumptions for the objects and their motion, followed by the discrete time dynamics model we adopt based on these assumptions. The object is placed on a horizontal planar surface such as tabletop. Thus its motion can be defined by $2D$ rigid body transformation $SE(2)$ within the horizontal plane while maintaining frictional contact with the underlying surface. We also assume the object motion is caused by external perturbation from a robot manipulator. Specifically, the manipulator grasps the object and apply force or torque on the object.  

\begin{table}[h]
	\centering 
	\caption{\small{Summary of Notations}}
	\label{tab:notation}
	\begin{tabular}{l l}
		\toprule
		Notation & Description  \\
		\midrule
		\makecell*[{{p{1.5cm}}}]{$\mb p_t$ $\mb v_t$ $\mb a_t$}        & \makecell*[{{p{6.5cm}}}]{Object pose, velocity and acceleration in $2D$ at time $t$, $\mb p = [p_x\ p_y\ p_w]$  where  $x,y$ denote translation  and $w$ denotes rotation. Same for velocity and acceleration} \\
		\makecell*[{{p{1.5cm}}}]{$M \ I_{cm} \ I_i$}  & \makecell*[{{p{6.5cm}}}]{Inertial mass and planar moment of inertia with respect to the center of mass or particle $i$} \\
		\makecell*[{{p{1.5cm}}}]{$\mb c$}  & \makecell*[{{p{6.5cm}}}]{Position of the center of mass $\mb c = [c_x \ c_y]$} \\
		\makecell*[{{p{1.5cm}}}]{$\mb p_i \ \mb v_i \ \hat{\mb v}_i$}  & \makecell*[{{p{6.5cm}}}]{Position, velocity and unit velocity of particle $i$ in $2D$, $\mb p_i = [p_{i,x}\ \ p_{i,y}]$ and $\mb v_i =[v_{i,x}\ \ v_{i,y}]$} \\
		\makecell*[{{p{1.5cm}}}]{$ n_p$}  & \makecell*[{{p{6.5cm}}}]{Number of particles in the model} \\
		\makecell*[{{p{1.5cm}}}]{$\mb m\ \ \bm\mu$}  & \makecell*[{{p{6.5cm}}}]{Vector of particle mass $m_i$ and friction coefficient $\mu_i$ in each group, with $dim(\mb m)=n_m$, $dim(\bm\mu)=n_\mu$} \\
		\makecell*[{{p{1.5cm}}}]{$\mb u_{i,t}$}  & \makecell*[{{p{6.5cm}}}]{$\mb u_{i,t} = \begin{bmatrix} u_{i,x} \ u_{i, y} \ u_{i,w} \end{bmatrix}_t$  the external perturbation applied on particle $i$} \\
		\bottomrule
	\end{tabular}    \emph{Notes:} bold uppercase letters for matrix, bold lowercase for column vector, normal ones are scalar
\end{table}

\subsection{Object Parameters}
\begin{figure}[ht!]
	\centering
	\includegraphics[width=0.4\textwidth]{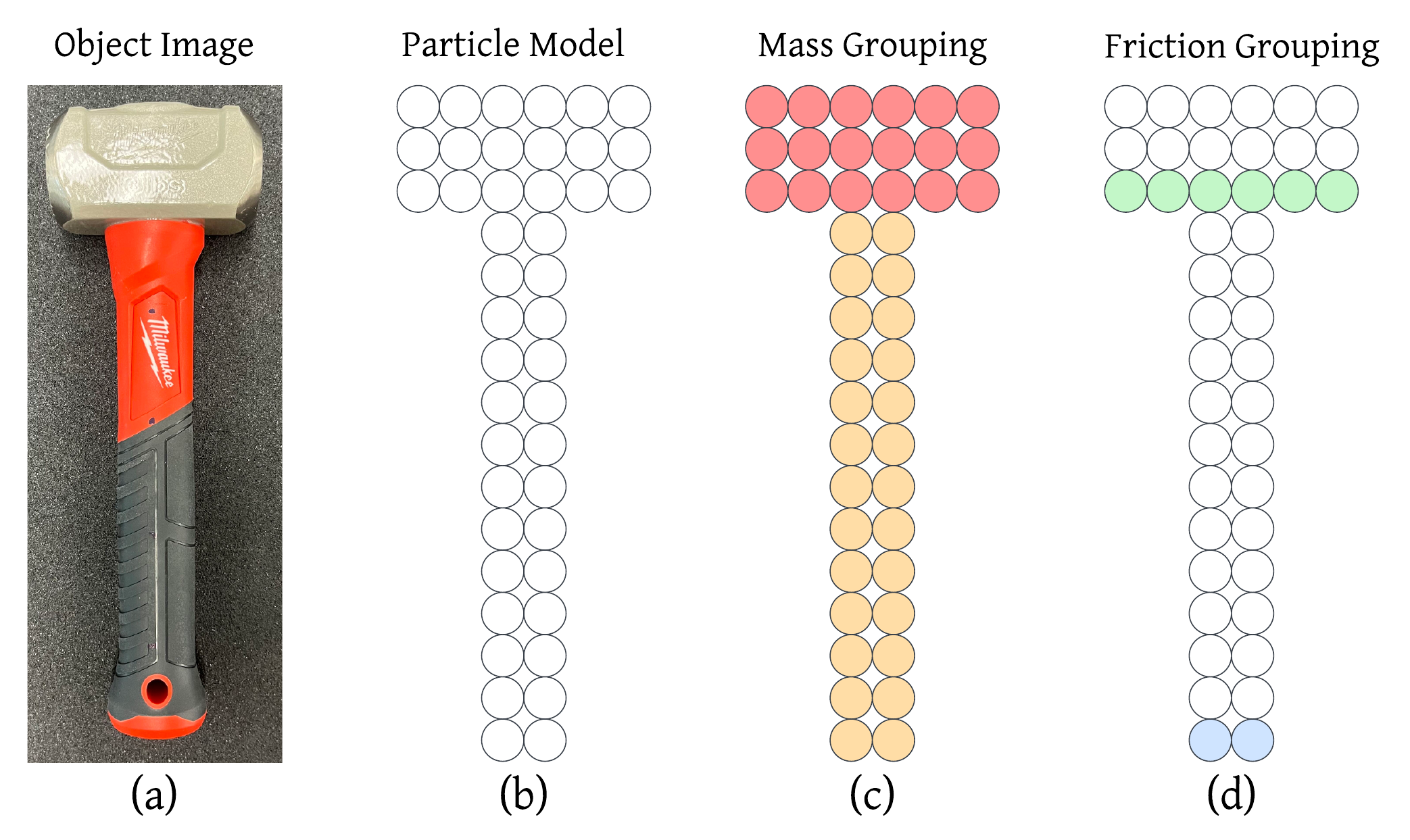}
	\caption{\small{An example of how we model the object properties. Based on the image (a) of the hammer, a particle model (b) is used to capture the object shape. We may also make assumptions about the mass grouping (c) where particles in the same group have the same mass (marked as red and orange). For the bottom contact (d), particles may have no contact (white) or in contact with different sliding friction parameter (green and blue).}}
	\label{fig:partition}
	\vspace{-25mm}
\end{figure}

There are two crucial aspects to consider when modeling the object properties: the mass distribution and the bottom contact surfaces. In Table~\ref{tab:notation}, we list the parameters we use to model the object properties and dynamics.  In Fig.~\ref{fig:partition}, we use a hammer as an example to illustrate the modeling process. First, a particle model is obtained by grid partition based on the 3D object model. Then the particles can be grouped into different regions in the mass group and contact surface group. In each region, the particles have the same parameter. For the simplicity of our model, we only consider sliding friction. We use $\mu mg$, where $\mu$ is friction coefficient, $m$ is the particle mass, and $g$ is gravity, as the magnitude for the sliding friction force on each particle. In Fig.~\ref{fig:partition}(d) only the colored regions (blue and green) are in contact. For particles not in contact with the bottom surface, $\mu$ will be $0$. 

In addition, we introduce the group mapping matrix $\mb G_m$ and $\mb G_\mu$ that maps the low dimension $\mb m$ and $\bm\mu$ to each particle. Specifically,
\begin{equation}
	\mb G_m = \begin{bmatrix} \mb e_1 & \hdots & \mb e_{n_p} \end{bmatrix} \ \text{and} \ \mb G_\mu = \begin{bmatrix} \mb g_1 & \hdots & \mb g_{n_p} \end{bmatrix}
\end{equation}
Here, $\mb e_i \in \mathbb{R}^{n_m}$ is a column vector of zeros except for the entry that corresponds to the index of the mass group. In the example of Fig.~\ref{fig:partition}, $\mb e_i$ can be either $[1,0]^T$ or $[0, 1]^T$. Similarly, $\mb g_i \in \mathbb{R}^{n_\mu}$ is computed the same way for particles with contact. But for non-contacting particles (white ones in Fig.~\ref{fig:partition}d), the corresponding column is all zeros.

\subsection{Object Dynamics}
Since the object is manipulated while being grasped, we consider one object in the system dynamics. This helps significantly simplify our model as it is no longer necessary to model the contact dynamics between multiple objects. 
According to Newton's second law, the translational acceleration of the object is determined by the force acting on it and its inertial mass. While the rotational acceleration is determined by the torque and its moment of inertia. The concept of inertial mass $M$ is similar to the moment of inertia from Newton's Law but the moment of inertia $I$ of a body also depends on the shape of the body, and the location and orientation of the axis of rotation relative to the body, making it a tensor. For rotation about a fixed axis, only a single scalar number from the moment of inertia tensor is required. In our 2D planar setting, the moment of inertia only depends on the point of the rotation center. 

As a result, we propose to use the following discrete update functions to describe the dynamics of the object from time $t$ to $t+1$:
\begin{align} 
	\mb p_{t+1} &= \mb p_{t} + \mb v_t \ dt  \label{eq:update1}\\
	\mb v_{t+1} &= \mb v_t + \mb a_t \ dt \label{eq:update2}
\end{align}

The acceleration $\mb a$ at each time step is a function of the object mass $M$, the moment of inertia $I_{cm}$, the center of mass $\mb c$, the combined effective force/torque from the friction force $\mb f_{i,f}$ and the external perturbation $\mb f_{i,u}$ on each particle $i$. The effective torque is a combination of the torque input and the cross product of the force input with the lever arm vector. In our work, we use $\mb u_i$ for the external perturbation applied on particle $i$ and assume the external perturbation is only applied to a single particle. Therefore on the other particles, both $\mb u_i$ and $\mb f_{i,u}$ become zero. 

Let $\otimes$ denote vector cross product. 
We may write $M$, $I_{cm}$, $\mb c$, $\mb f_{i,f}$ and $\mb f_{i,u}$ as follows:
\begin{align}
	M &= \sum \mb G^T \mb m \label{eq:mass}\\
	\mb c  &= \sum_i (\mb e_i^T \mb m) \mb p_i / M  \label{eq:com}\\
	I_{cm} &= \sum_i (\mb e_i^T \mb m) \ \lVert\mb p_i - \mb c\lVert^2  \label{eq:inertia} \\ 
	\mb f_{i, u} &= \begin{bmatrix}
		u_{i,x} \\ u_{i,y} \\ u_{i,w} + \begin{bmatrix} u_{i,x} \\ u_{i, y} \end{bmatrix} \otimes 
		(\mb p_i - \mb c)
	\end{bmatrix} \\
	\mb f_{i, f} &= - (\mb g_i^T \bm\mu) (\mb e_i^T \mb m) g \begin{bmatrix}
		\hat{v}_{i, x} \\ \hat{v}_{i, y} \\ \hat{\mb v}_i \otimes 
		\left(\mb p_i - \mb c\right)
	\end{bmatrix}\label{eq: fif}
\end{align}

Then the acceleration vector $\mb a$ can be written as:
\begin{equation} \label{eq:dv}
	\mb a = \mb M \sum_i (\mb f_{i, u} + \mb f_{i, f} ) \
	\text{where, } \mb M =  \begin{bmatrix} M &&\\&M&\\&&I_{cm} \end{bmatrix}^{-1}
\end{equation}

\begin{figure}[ht!]
	\centering
	\includegraphics[width=0.4\textwidth]{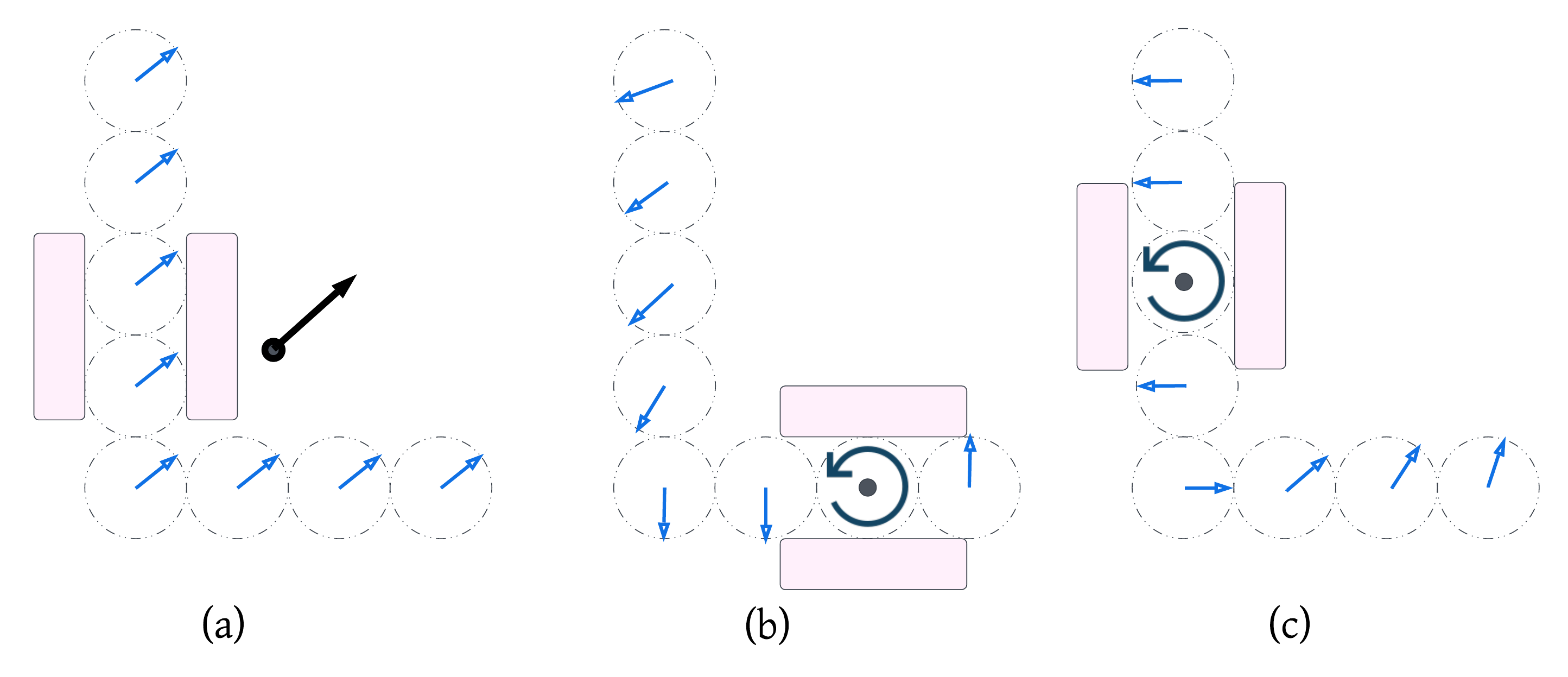}
	\caption{\small{An example of the robot actions: "grasp-and-slide" and "grasp-and-rotate". The pink rectangles illustrate the two fingers of the robot gripper. The black arrows and blue arrows correspond to object velocity and particle velocity. (a) In translational motion, all particles have the same velocity as the object. (b,c) In rotational motion, particle velocities are dependent on the rotation axis.}}
	\label{fig:trajectory}
	\vspace{-4mm}
\end{figure}

\subsection{Robot Actions} \label{sec:actions}
The $SE(2)$ motion of the object consists of translation and rotation and we classify the robot actions accordingly. In Fig.~\ref{fig:trajectory}, we use an L-shaped object to illustrate the idea. The object with 10 particles is grasped by the parallel jaw gripper (pink rectangles for two fingers) at different locations. The black arrows show the object velocity, where Fig.~\ref{fig:trajectory}a moves towards upper right, and Fig.~\ref{fig:trajectory}b and Fig.~\ref{fig:trajectory}c rotate about the particle centers of rotation. It is worth noticing that, in Fig.~\ref{fig:trajectory}a, all particle velocities (blue arrows) are the same as the object velocity. But in Fig.~\ref{fig:trajectory}b, and Fig.~\ref{fig:trajectory}c, the particle velocity depends on the rotation center and the velocity of the particle at the rotation center is zero. Thus, we name the robot actions as ``grasp-and-slide" or ``grasp-and-rotate" for pure translational or pure rotational motions.

\section{Problem Formulation}
We aim to estimate the 2D mass distribution of an object by applying mechanical perturbation and observing the motion of the object.
Given an object on a 2D plane with its grid partition and mass/friction grouping, a robot manipulator with a gripper, and a force/torque sensor that can measure the mechanical perturbation applied to the object, 
the goal is to estimate the object mass distribution $m_i$ by measuring the tactile feedback and the object trajectory while the robot grasps and moves the object around.

Let $\Phi$ be the object forward dynamics function and $\mb u_{i,t}$ be the external perturbation on particle $i$, we may write $\Phi$ as:
\begin{equation}
	\mb v_{t+1} = \Phi \left( \mb p_{t}, \mb v_{t}, \mb m, \bm{\mu} , \mb u_{i,t}\right)
\end{equation}
we want to find the optimal $\mb m$ and $\bm\mu$ that minimize the error of the object trajectory predicted by the dynamics model and the one observed as ground truth. So our problem can be formulated as computing:
\begin{equation} \label{eq:problem}
	\left(\mb m^*, \bm\mu^*\right) = 
	\argmin_{\mb m, \bm\mu} \sum_{t=0}^T\lVert \Phi\left(\mb p_t, \mb v_t, \mb m, \bm\mu, \mb u_{i,t}  \right) -  \mb v_{t+1}\lVert^2 
\end{equation}

\section{Our Approach}
Our approach to solving the mass distribution estimation problem is built on top of the analytical formulation and the dependencies of the dynamics model parameters on the external perturbations as well as the object trajectory. In this section, we first introduce a new set of derived parameters called ``Hidden States". Then we talk about the algorithms we designed to estimate the Hidden States parameters.

\subsection{Hidden States}
The discrete-time update equations of system dynamics show that the model has high non-linearity with respect to the model parameters $m_i$. It is challenging to apply traditional nonlinear system identification methods such as the Volterra series methods because the data requirement scales exponentially with respect to the order of polynomials. Additionally, the friction parameter $\mu_i$ is also unknown and $m_i$ is multiplied by $\mu_i$ in Eq.~\ref{eq: fif}, further complicating the problem. 

However, by introducing the Hidden States $\mb H$ as follows, we will show how to reduce the model complexity to linear or quadratic models and observe the Hidden State parameters sequentially:
\begin{equation}
	\mb H = \begin{bmatrix}
		M & I_{cm} & \mb c & \mb s
	\end{bmatrix}
\end{equation}
where, $\mb s$ is a vector for the set of $\mu  m g$ for all particles with contact, and we may adopt the group mapping matrix $\mb G_s =\begin{bmatrix}\mb h_1 \ ... \ \mb h_{n_p}\end{bmatrix}$ for $\mb s$, similar to $\mb m$ and $\bm\mu$.

The Hidden States $\mb H$ consist two types of parameters:
\begin{itemize}
	\item object-level parameters including the inertial mass $M$, moment of inertia $I_{cm}$ and center of mass $\mb c$
	\item particle-level parameters $s_i$ due to the friction interaction between the bottom surface with the table
\end{itemize}
Our approach is to first estimate the object level parameters in the Hidden States followed by the particle level parameters. Then $m_i$ can be derived arithmetically based on Eq.~\ref{eq:mass}-\ref{eq:inertia}. 

\vspace{-3mm}
\begin{algorithm}[h]
	\caption{Multi-stage Hidden States Estimation}
	\label{alg:hidden_state}
	\SetAlgoLined
	Initialize with uniform mass $\mb m$ and zero friction $\bm\mu$\\
	Compute robot action set $S$ \\
	Sample $k$ grasp-and-rotate actions from $S$ to $S_1$ with $k\geq 3$\\
	\For{each action $ \in S_1$}
	{
		Grasp the object at $\mb p_i$ \\
		Apply various torque $u_{i,w}$ and record $a_w$\\
		Compute $I_i$ by linear fitting using Eq.\ref{eq:dvp}
	}
	Solve Eq.~\ref{eq:problem_com_inertia} for $\mb c$ and $I_{cm}$\\
	Sample $n_s / 3$ grasp-and-rotate actions from $S$ to $S_2$\\
	Compute $rank(Q)$ \\
	\While{$rank(Q) < n_s$}
	{
		Add one more sample to $S_2$ \\
		Compute $rank(Q)$ \\
	}
	\While{not timeout nor convergence of $\mb s$}
	{
		\For{each action $\in S_2$}
		{
			Compute forward dynamics
		}
		Compute $\mathcal{L}(\mb s)$ and $\frac{\partial \mathcal{L}}{\partial \mb s}$\\
		$s = s - \alpha_{rate} * \frac{\partial \mathcal{L}}{\partial \mb s}$
	}
	Solve for $\mb m$ from Hidden States $\mb H$
\end{algorithm}

\subsection{Parameter Estimation}
In the object-level parameters, one significant difference between the inertial mass and the moment of inertia is that gravitational mass is equivalent to inertial mass. So the mass can be determined by weighing rather than from Newton’s second law. The moment of inertia, however, can only be measured by observing the force/torque and angular acceleration. So the inertial mass $M$ can be directly observed by weighing the object without having the robot move it. For the rest of the object level parameters, i.e. $[c_x \ c_y]$ and $I_{cm}$, they can be jointly estimated by first observing $I_j$.

During a ``grasp-and-rotate" action around the particle $j$. The angular acceleration of the object $a_w$ can be described by replacing $\mb c$ with  $\mb p_j$ in Eq.~\ref{eq:inertia} and Eq.~\ref{eq: fif}.
Therefore, $a_w$ in Eq.~\ref{eq:dv} can be rewritten as:
\begin{equation*} 
	a _w = I_j^{-1} \left(  u_{j,w}-  \sum_i   (\mb h_i^T \mb s) 
	\hat{\mb v}_i \otimes 
	\left(\mb p_i - \mb p_j \right)
	\ \right)
\end{equation*}
\begin{equation}\label{eq:dvp} 
	\text{where, }   I_j = \sum_i (\mb e_i^T \mb m) \ \lVert\mb p_i - \mb p_j\lVert^2 
\end{equation}
From Eq.~\ref{eq:dvp}, we can see that the output $a_w$ is linearly dependent on the input torque $u_{j,w}$. Therefore, $I_j$ can be observed by applying different torque in such grasp-and-rotate motion and observing the angular acceleration of the object.
Furthermore, we know from the parallel axis theorem that $I_j$ is a function of $I_{cm} \ \text{and}\ \mb c$:
\begin{equation}
	I_j = I_{cm} + M  \lVert\mb p_j - \mb c\lVert^2 
\end{equation}

Let $S$ be the set of the index $j$ for which $I_j$ is observed, the estimation of $\mb c$ and $I_{cm}$ can be formulated as computing:
\begin{equation} \label{eq:problem_com_inertia}
	\left(\mb c^*, I_{cm}^*\right) = 
	\argmin_{\mb c, I_{cm}} \sum_{j \in S}\lVert I_{cm} + M  \lVert\mb p_j - \mb c\lVert^2 - I_j\lVert^2 
\end{equation}

Once we compute the object-level parameters, we can treat them as known variables in Eq.~\ref{eq:dv}. As a result, both $\mb v_{t+1}$ and $\mb a_t$ in Eq.~\ref{eq:update2} are linear with respect to $\mb s$, and 
Eq.~\ref{eq:update2} can be rewritten in vector product form as: 
\begin{align} \label{eq:dvs} 
	\mb v_{t+1} & = \mb v_t + \mb A_t \ \mb G_s^T \ \mb s + \mb B_t \quad \text{where, } \nonumber\\
	\mb A & = dt \ \mb M \begin{bmatrix} \hat{v}_{0, x} &  ... & \hat{v}_{i, x} & ...  \\ 
		\hat{v}_{0, y}  & ... & \hat{v}_{i, y}  & ...\\ 
		\hat{\mb v}_0 \otimes \left(\mb p_0 - \mb c\right) & ... & \hat{\mb v}_i \otimes \left(\mb p_i - \mb c\right) & ...
	\end{bmatrix} \nonumber \\
	\mb B & = dt \ \mb M \sum_i \mb f_{i,u} 
\end{align}

Therefore, we could adopt the least square method to estimate $s$ by populating observations from different trajectories of the object. If we use superscript $k$ to denote the parameters in the $k^{th}$ trajectory, the problem in Eq.~\ref{eq:problem} reduces to:
\begin{align} \label{eq:problem_si}
	\mb s^* &=  \argmin_{\mb s} \mathcal{L}(\mb s) \quad \text{where, } \nonumber \\
	\mathcal{L}(\mb s) &= \sum_{t=0}^T
	\lVert   \begin{bmatrix} \mb A^0 \\ \vdots \\ \mb A^k \\ \vdots \end{bmatrix}_t  \mb G_s^T \mb s + 
	\begin{bmatrix} \mb B^0 \\ \vdots \\ \mb B^k \\ \vdots \end{bmatrix}_t +  
	\begin{bmatrix} \mb v^0 \\ \vdots \\ \mb v^k \\ \vdots  \end{bmatrix}_t - 
	\begin{bmatrix} \mb v^0 \\ \vdots \\ \mb v^k \\ \vdots  \end{bmatrix}_{t+1} \lVert^2 
\end{align}

To be able to solve for Eq.~\ref{eq:problem_si}, the matrix $\mb Q = [\mb A^0 ... \mb A^k ...]^T  \mb G_s^T \in \mathbb{R}^{3k \times n_s}$  must be full column rank, i.e. $rank(\mb Q)=n_s$.
Based on Section.~\ref{sec:actions}, we only need to analyze two special cases of the trajectories, which corresponds to two special robot actions: constant speed ``grasp-and-slide" and constant speed ``grasp-and-rotate". 

For ``grasp-and-slide" actions, all particles have the same velocity as the rigid object: $[v_{i,x} \ v_{i,y}] = [v_x \ v_y]^T$
This means the first two rows of $\mb A$ become: $  M^{-1} \hat{\mb v} \begin{bmatrix} 1 &  ... & 1 & ... \end{bmatrix}$.
As a result, the first two rows of all $\mb A^k$ in $\mb Q$ are linearly dependent on each other, giving $k+1$ as the upper bound to $rank(\mb Q)$. 

However, for ``grasp-and-rotate" actions with rotation center around particle $j$, the first two rows of $\mb A$ become: $
M^{-1}  \begin{bmatrix}\hat{\mb v}_{0} &  ... & \hat{\mb v}_{j-1} & \mb 0  & \hat{\mb v}_{j+1} & ... \end{bmatrix}$, where only the $j^{th}$ entry is zero.
This means the vectors from two different  ``grasp-and-rotate" actions are guaranteed to be linearly independent, making $\mb Q$ less degenerated compared to the $\mb Q$ constructed from ``grasp-and-slide" actions.
Therefore, we adopt a random sampling method with higher priorities for ``grasp-and-rotate" actions to observe the particle-level parameters.

Algorithm.~\ref{alg:hidden_state} summarize the multi-stage estimation process to actively estimate the parameters in the Hidden States.

\subsection{Differentiable Physics}
In order to efficiently solve the problem in Eq.~\ref{eq:problem_si}, we propose to use a differentiable physics engine for the forward dynamics with the Hidden States. The jacobian $\frac{\partial \mathcal{L}}{\partial \mb s}$ in Algorithm.~\ref{alg:hidden_state} can be obtained through auto-differentiation, and then a gradient descent on the loss function in Eq.~\ref{eq:problem_si} can be performed. In our work, instead of the popular auto-differentiation frameworks like PyTorch or TensorFlow, we adopted Taichi Lang \cite{hu2019difftaichi} to implement a computationally efficient differentiable physics model.

\section{Evaluation}

In this section, we first introduce our real robot setup and block object design followed by our evaluation metrics for the algorithms. Then we continue our section with the baseline methods used for comparisons and evaluation.

\begin{figure}[ht!]
	\centering
	\vspace{-2mm}
	\includegraphics[width=0.37\textwidth]{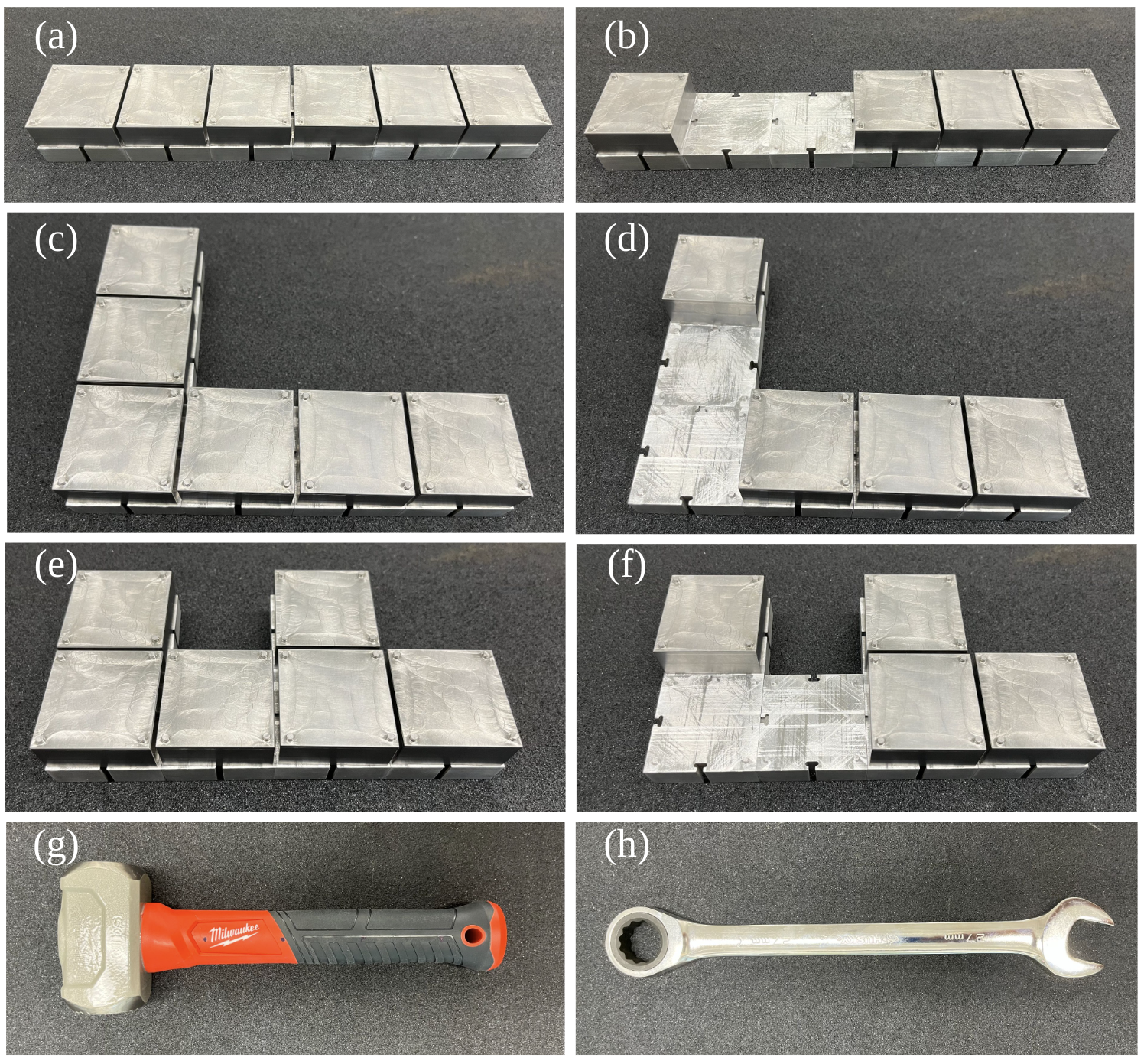}
	\caption{\small{The objects we used for our experiments. (a-f) shows different configurations of the block object. (g) a 3 lb hammer (f) a wrench }}
	\label{fig:object}
\end{figure}

\subsection{Real Robot Implementation Details}
For the real robot setup, we use the tabletop manipulator setting shown in Fig.~\ref{fig:concept}. We mount the Robotiq FT-300S 6-axis force/torque sensor on the wrist of a Kinova Gen 3 7-DOF manipulator. 
The force/torque feedback is measured at 100 Hz while the manipulator performs servoing to follow the trajectories defined by the selected actions. 
For ``grasp-and-slide" actions, the robot end-effector moves at a constant speed of $5~cm/s$ along a straight line for $4$ seconds. 
For ``grasp-and-rotate" actions, the robot rotates its end-effector at a constant angular speed of $10~deg/s$ for $18$ seconds.
Instead of taking the measured raw tactile data directly, we take advantage of the physics behind the constant speed robot actions we designed. In constant-speed ``grasp-and-rotate" actions, the object has no change in the rotational kinetic energy and the center of mass is circulating with respect to the rotation axis. Therefore the measured tactile feedback should have constant torque and sinusoidal force. In constant-speed ``grasp-and-rotate" actions, the object has no change in both the translation and rotation kinetic energy, so the measured tactile feedback should have constant force and torque. We utilize such properties to pre-process and filter our data and ground truth trajectories, which allowed us to reduce the noise from the data collection process.

As for the objects for the experiments, we designed and fabricated a re-configurable block object set besides regular objects such as a hammer and a wrench. The block object is made from blocks of aluminum or steel and can change shape or mass distribution. This design allows us to capture the ground truth mass distribution for better evaluation. Fig.~\ref{fig:object} shows the objects we used for the experiments. We name the block objects in Fig.~\ref{fig:object} (a-f) as ``I1", ``I2", ``L1", ``L2", ``F1", and ``F2" accordingly.

\begin{table}[h]
	\centering
	\caption{\smaller{Normalized Absolute Difference of Mass Distribution  ($\%$) of each block object (column) and each approach (row)}}
	\label{tab:nad}
	\begin{adjustbox}{width=\linewidth}
		\begin{tabular}{l c c c c c c} 
			\toprule
			{} &  I1 & I2 & L1 & L2 & F1 & F2 \\
			\midrule		   
			Our Approach                &  \textbf{7.4}  &  \textbf{9.2} &  \textbf{10.1} &  \textbf{7.3} &  \textbf{8.5} & \textbf{14.2} \\
			Random Search               &  56.5 & 32.6 & 31.1 & 46.4 & 24.2 & 42.9\\
			Weighted Sampling Search    &  30.1 & 25.1 & 44.4 & 32.9 & 27.9 & 25.0 \\
			Explicit State              &  18.2 & 22.7 & 35.0 & 29.4 & 22.8 & 34.9 \\
			Diff-LCP                    &  44.2 & 51.7 & 32.2 & 50.4 & 36.3 & 68.5 \\
			\bottomrule
		\end{tabular}
	\end{adjustbox}
\end{table}

\subsection{Evaluation Metrics}
Although we could capture the ground truth mass distribution for the block objects, it is difficult to do so for the hammer and the wrench. Therefore we propose two evaluation metrics to evaluate the estimated mass distribution.

For the estimation results from the block objects, let $\mb m$ and $\mb m_g$ be the estimated and ground truth mass distributions. The \textit{Normalized Absolute Difference} (NAD) $\eta_1$ is computed by summing the absolute error for each particle and then normalizing by the total mass:
\begin{equation}
	\eta_1 = \frac{\sum\mb G_m^T \ \lvert \mb m - \mb m_g \rvert}{\sum \mb G_m^T  \mb m_g }
\end{equation}

For further evaluation, we also run the forward simulation with the estimated results to compare against the ground truth trajectories observed on the real robot setup. Instead of computing the difference for all time steps along the trajectories, we only compare the mean particle distance at the final time step. Let $\mb p_i$ and $\mb p_{i,g}$ be the simulated and ground truth particle position at the last time step, we define the \textit{Mean Particle Distance} (MPD) $\eta_2$ as:
\begin{equation}
	\eta_2 = \frac{\sum_i \lVert \mb p_i - \mb p_{i,g} \rVert}{n_p}
\end{equation}

\subsection{Baseline Methods}
Next, we go through the methods we used as baselines to compare against our approach.

\textbf{Random Search} \cite{random_search}randomly samples $\mb m$ and $\bm\mu$ repeatedly and runs the forward simulation to search for the sample that returns the minimal loss $\mathcal{L}$.

\textbf{Weighted Sampling Search}\cite{weighted_sampling} is initialized with grid sampling method for $\mb m$ and $\bm\mu$. And then iterative by re-sampling from a Gaussian distribution around the best sample in the last iteration. The sampling deviation is decreased over time to focus on the most promising sampling region.

\textbf{Explicit State} uses the same analytical formulation for the dynamics as our approach. But instead of first estimating the Hidden States, this method directly computes the gradient of $\mathcal{L}$ with respect to $\mb m$ and $\bm\mu$ and run gradient descent. 

\textbf{Diff-LCP} \cite{song2020identifying} adopts the same particle-based object model and uses a differentiable physics engine based on LCP as the forward simulation for the dynamics. Diff-LCP requires a smaller update time step so we do linear interpolation on the tactile feedback data before feeding them into the forward simulation. Diff-LCP also computes the gradient $\mathcal{L}$ with respect to $\mb m$ and $\bm\mu$ and runs gradient descent to optimize the estimation. 

\begin{figure}[ht!]
	\centering
	\includegraphics[width=0.38\textwidth]{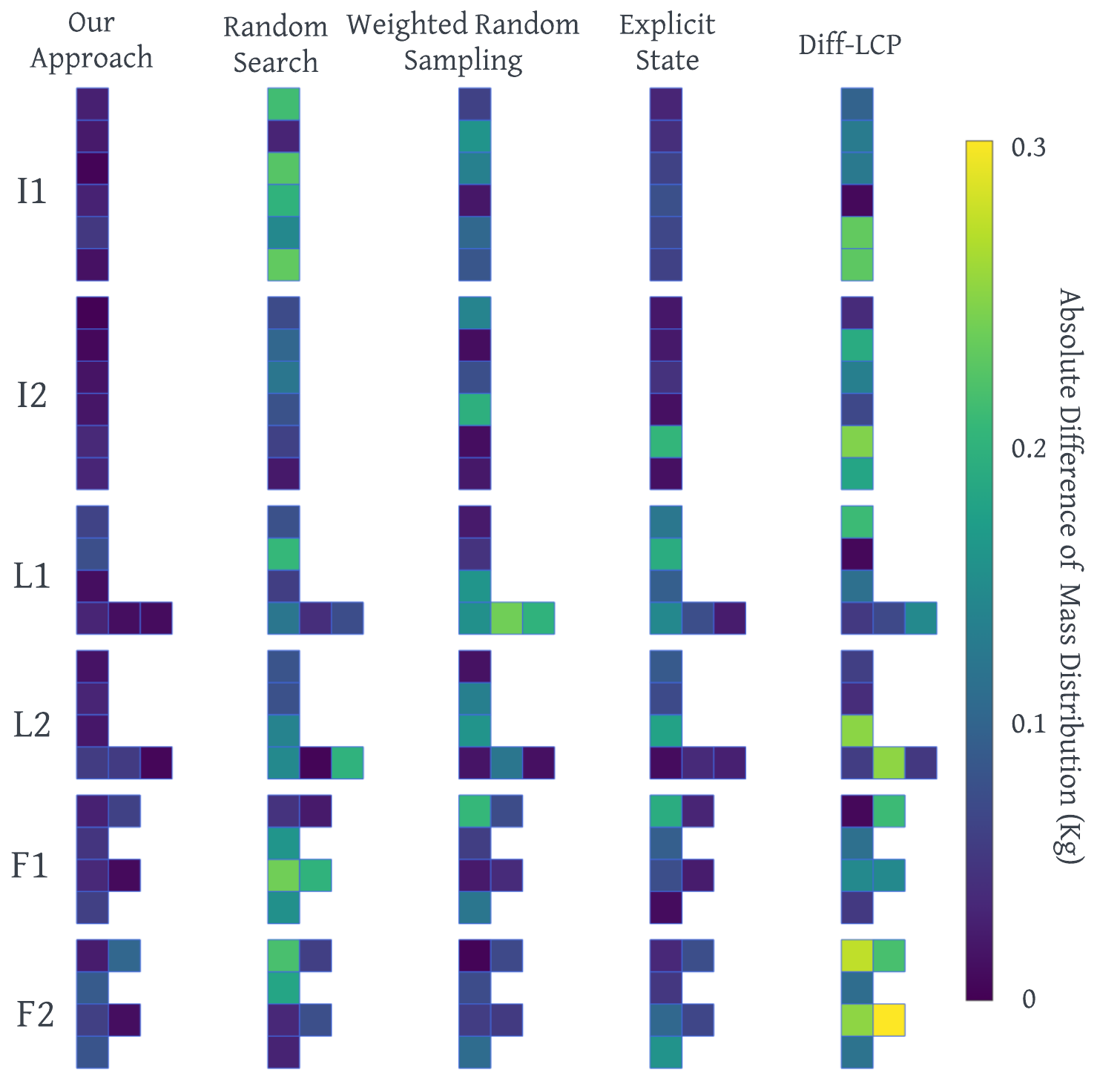}
	\caption{\small{Heatmap of the Absolute Difference of mass (in kilograms) for each block in the block objects. Each column corresponds to different methods and each row corresponds to different block object configurations.}}
	\label{fig:absolute_diff}
\end{figure}

\subsection{Results}

\begin{table}[h]
	\centering
	\caption{\smaller{Mean Particle Distance from Forward Simulation ($cm$) of each block object (column) and each approach (row)}}
	\label{tab:mpd}
	\begin{adjustbox}{width=\linewidth}
		\begin{tabular}{l c c c c c c c c} 
			\toprule
			{} &  I1 & I2 & L1 & L2 & F1 & F2 & hammer & wrench \\
			\midrule		   
			Our Approach                &  \textbf{1.2}  &  \textbf{1.5} &  \textbf{1.1} & \textbf{0.8} & \textbf{1.6} & \textbf{2.2} & \textbf{1.9} & \textbf{1.3} \\
			Random Search               &  9.1  & 13.8 &  7.6 & 8.2 & 9.3 & 9.6 & 3.4 & 6.1 \\
			Weighted Sampling Search    &  5.4  &  9.8 &  5.8 & 5.9 & 4.6 & 7.5 & 4.0 & 4.2 \\
			Explicit State              &  1.7  &  2.6 &  1.8 & 1.5 & 2.9 & 3.2 & 3.3 & 3.6 \\
			Diff-LCP                    &  6.2  &  5.3 &  6.8 & 4.8 & 3.5 & 7.3 & 4.6 & 3.7 \\
			\bottomrule
		\end{tabular}
	\end{adjustbox}
\end{table}

Since we do not know the ground truth mass distribution for the hammer and wrench, we only compare NAD for the block object set. For MPD, we could compare across all the objects. We run our approach with 500 gradient steps. For fair comparisons, we also run each of the baseline methods for 500 iterations. Table~\ref{tab:nad} shows the NAD of the mass distribution after 500 iterations for each object and each method. All gradient-based methods are tested with various optimizers 
such as Adam \cite{kingma2014adam} and RMSProp \cite{rmsprop}, 
and the best result is picked for evaluation. To evaluate the forward predictions on the estimated mass distribution, we take the ``grasp-and-rotate" actions that are not sampled during the data collection stage. Table~\ref{tab:mpd} shows the MPD at the final time step for each object and each method.

Our approach outperforms all the baselines in both evaluation metrics. Both Random Search and Weighted Sampling Search perform poorly. Specifically, Random Search suffers from the high dimensional sampling space from the block objects. Weighted Sampling Search shows high sensitivity to the sampling deviation decay rate. The reason for the sensitivity could be aligned with what prevents the Explicit State method perform well. In the Explicit State method, we always observe significantly slow updates after the initial few iterations. It is likely that the estimation is trapped by ``local minimum" or ``valley" in the joint space of mass and friction. As a comparison, our approach avoids the estimation within the joint space by first estimating the Hidden States sequentially and computing the mass distribution analytically. Diff-LCP  adopts a different dynamics model from all the methods above. The results show that although it could reduce the reality gap (MPD) to a certain extent, the forward model is physically inaccurate, making the estimation results off by a large margin.

\section{Conclusion}
We presented a method to actively estimate the planar mass distribution of a rigid object through tactile feedback during object manipulation. We combined a sequential estimation strategy with a set of robot action selection rules based on the analytical formulation of a discrete-time dynamics model. Our evaluation results on objects with various shapes and mass distributions show that our approach can solve this challenging task with high accuracy. 
Exciting future research directions include extensions to more complex objects such as articulated ones and analysis of action selection rules that will guarantee the uniqueness of the estimation result.

\bibliographystyle{IEEEtran}
\bibliography{mass.bib}

\end{document}